\documentclass{IOS-Book-Article}

\usepackage{mathptmx}
\usepackage{soul}\setuldepth{article}
\usepackage{ amssymb }
\usepackage{ amsmath }
\usepackage{blkarray,booktabs, bigstrut}
\usepackage{graphicx}
\usepackage{tikz}

\usetikzlibrary{shadows.blur}
\usetikzlibrary{shapes.geometric}
\usetikzlibrary{positioning}
\usetikzlibrary{matrix}
\usetikzlibrary{arrows}
\usetikzlibrary{backgrounds}
\usetikzlibrary{shapes.multipart}
\usetikzlibrary{decorations.pathreplacing}
\newcommand{\body}{\textbf{B}}
\newcommand{\hornrule}{\textbf{B} \Rightarrow H}
\newcommand{\kg}{\mathcal{K}}

\newcounter{todocounter}

\def\hb{\hbox to 11.5 cm{}}

\begin{document}

\pagestyle{headings}
\def\thepage{}
\begin{frontmatter}              

\title{Neurosymbolic Methods for Rule Mining}

\markboth{}{August 2024\hb}

\author[A]{\fnms{Agnieszka} \snm{Ławrynowicz}\orcid{0000-0002-2442-345X}%
\thanks{Corresponding Author: Author Name, contact details.}},
\author[B]{\fnms{Luis} \snm{Galárraga}\orcid{0000-0002-0241-5379}}
\author[C]{\fnms{Mehwish} \snm{Alam}\orcid{0000-0002-7867-6612}}
\author[C]{\fnms{Bérénice} \snm{Jaulmes}}
\author[D]{\fnms{Václav} \snm{Zeman}}
and
\author[D]{\fnms{Tomáš} \snm{Kliegr}\orcid{
0000-0002-7261-0380}}

\runningauthor{B.P. Manager et al.}
\address[A]{Poznan University of Technology, Poland}
\address[B]{INRIA/IRISA, Rennes, France}
\address[C]{Télécom Paris, Institut Polytechnique de Paris, France}
\address[D]{Prague University of Economics and Business, Czech Republic}

\begin{abstract}
In this chapter, we address the problem of rule mining, beginning with essential background information, including measures of rule quality. We then explore various rule mining methodologies, categorized into three groups: inductive logic programming, path sampling and generalization, and linear programming. Following this, we delve into neurosymbolic methods, covering topics such as the integration of deep learning with rules, the use of embeddings for rule learning, and the application of large language models in rule learning.
\end{abstract}

\end{frontmatter}
\markboth{June 2024\hb}{June 2024\hb}

\section{Introduction}
The schema of knowledge graphs (KGs) can be represented using ontological axioms and/or rules.
Rules can be used for explainable inference for tasks such as link prediction or fact checking~\cite{boschin_combining_2022}.

Hovewer, formulating rules manually is demanding in practice. 
For that reason, automatic rule learning approaches have attracted attention.

Wu et al.~in their survey~\cite{wu_rule_2023} distinguish three major groups of rule learning methods: Inductive Logic Programming-based, statistical path generalisation and neuro-symbolic. 
In this chapter, to introduce the topic of rule mining and provide necessary backround, we discuss each of these groups and give a more detailed example of algorithms within each group paying most attention to neuro-symbolic ones. 
We also discuss the topic of using Large Language Models (LLMs) for rule mining.

\section{Background}


\begin{figure}
	  	\begin{tikzpicture}
	\tikzstyle{mynode}=[ellipse, draw,
	top color=gray!5, bottom color=gray!20, 
	shade, minimum width=1.5cm]
	\tikzstyle{mynodet}=[ellipse, draw, 
	top color=blue!5, bottom color=blue!20, 
	shade, minimum width=1.5cm]
	\tikzstyle{mynodes}=[ellipse, draw, 
	top color=green!20!gray!50, bottom color=green!50!black!80, 
	shade, minimum width=1.5cm]
	\tikzstyle{elabel}=[fill=white, align=center, font={\footnotesize}]
    \node(amv) at (5.3,0) [mynode] {A. Merkel};
	\node(am) at (0,2.5) [mynode] {U.v.d. Leyen};
	\node(em) at (0,-2.5) [mynode] {E. Macron};
	\node(female) at (-3.8,0.5) [mynode] {female};
	\node(eu) at (-2,0) [mynode] {EU};
	\node(english) at (-4,2.7) [mynode] {English};		
	\node(deutsch) at (3.8,2.7) [mynode] {German};		
	\node(francais) at (3.8,-2.7) [mynode] {French};		
	\node(germany) at (1.1,0.5) [mynode] {Germany};
	\node(france) at (1.3,-0.5) [mynode] {France};		
	\node(male) at (-3.8,-0.5) [mynode] {male};
 	\draw[-latex, thick, bend right] (amv)
	edge node[elabel] {nationality} (germany.east);
  	\draw[-latex, thick, bend left] (amv)
	edge node[elabel] {birthCountry} (germany.east);
	\draw[-latex, thick, bend right] (am)
	edge node[elabel] {gender} (female.north);
	\draw[-latex, thick, bend right] (am)
	edge node[elabel] {worksFor} (eu.west);
	\draw[-latex, thick, bend left] (am)	

    edge node[elabel] {nationality } (germany.north);
	\draw[-latex, thick, bend right] (am)	
 edge node[elabel] {birthCountry } (germany.west);
	\draw[-latex, thick, bend right] (am)	
 edge node[elabel] {speaks} (english.north);		
	\draw[-latex, thick, bend right] (em)
 edge node[elabel] {nationality} (france.south);
		\draw[-latex, thick, bend left] (em)
 edge node[elabel] {birthCountry} (france.west);
	\draw[-latex, thick, bend left] (em)
	edge node[elabel] {gender} (male.south);
\draw[ -latex, thick, bend left] (am)
 edge node[elabel] {speaks} (deutsch.west);

 edge node[elabel] {nativeLang} (deutsch.west);

 \draw[ -latex, thick, bend left] (am)
 edge node[elabel] {speaks} (deutsch.west);

 edge node[elabel] {nativeLang} (deutsch.south);
 
	\draw[-latex, thick, bend right] (em)
	edge node[elabel] {speaks} (francais.west);
		\draw[-latex, thick, bend left] (em)
	edge node[elabel] {worksFor} (eu.south);
	\draw[-latex, thick, bend left] (france)
 edge node[elabel] {officialLang} (francais.north);
		\draw[-latex, thick, bend right] (germany)
 edge node[elabel] {officialLang} (deutsch.south);
\end{tikzpicture}
\caption{Sample knowledge graph.}
\label{fig:sample_kg}
\end{figure}
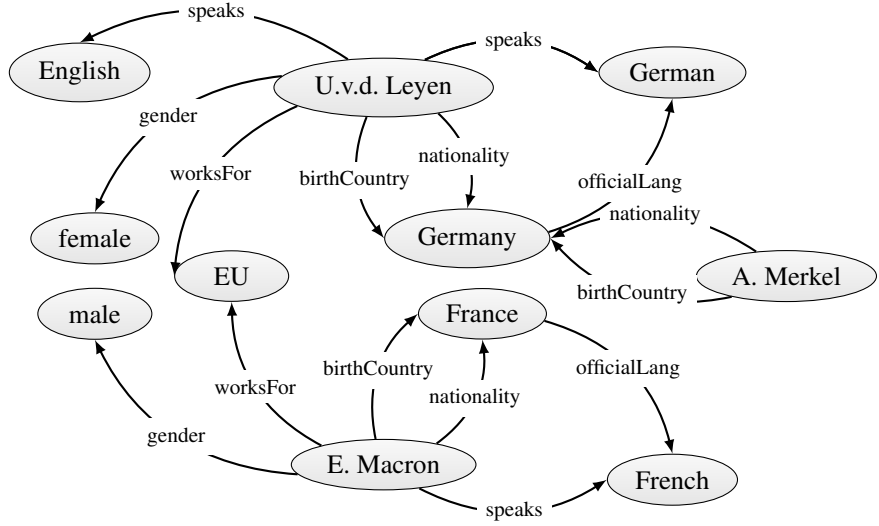

\paragraph{Rules} 
\label{sec:rules}
A (Horn) rule is an expression of the form 
\begin{equation}
q_1(z_0, z_1) \wedge...\wedge q_n(z_{n-1}, z_{n}) \Rightarrow p(x, y)    \label{eq:1}
\end{equation}

\noindent where $p(z, y)$ and each term $q_i(z_{i-1}, z_i)$ is an atom, that is, a KG fact such that at least one of its terms is a variable $v \in \mathcal{V}$. In the remainder of this chapter we denote variables $v \in \mathcal{V}$ by lowercase letters, and constants by capitalized names, e.g., \emph{Germany} $\in \mathcal{E}$, where $\mathcal{E}$ is a set of entities. The left-hand side part of the rule is a logical conjunction of atoms that we call the \emph{body} or \emph{antecedent} of the rule, denoted by $\body$, whereas the right-hand side atom is called the \emph{head} or the \emph{succedent} of the rule, denoted by $H$. We say two atoms are connected if they share at least one argument. A conjunction of atoms or a rule is connected if every atom is transitively connected to every other atom. For instance the rule $\textit{nationality}(x, z_1) \wedge \textit{officialLang}(z_1, y) \Rightarrow \textit{nationality}(x, y)$ is connected, whereas the rule $\textit{speaks}(x, z_1) \wedge \textit{officialLang}(z_2, z_3) \Rightarrow \textit{speaks}(x, y)$ is not because the second atom is not reachable from any other atom. Applications on KGs usually require connected rules.   

We say a rule is \emph{safe} if the head variables are present in the rule's body. For example, the rule 
$\textit{birthCountry}(x, z_1) \wedge \textit{nationality}(x, z_2) \wedge \textit{officialLang}(z_2, z_3) \Rightarrow \textit{speaks}(x, y)$
is not safe because variable $y$ is absent in the antecedent, which actually means that $y$ is existentially quantified. In other words, this rule could be interpreted as $\forall x\; \exists y, z_1, z_2, z_3 : \textit{birthCountry}(x, z_1) \wedge \textit{nationality}(x, z_2) \wedge \textit{officialLang}(z_2, z_3) \Rightarrow \textit{speaks}(x, y)$. Safeness ensures that the rule makes concrete predictions. In our example, the rule states that if someone, with known birth country is citizen of a country with an official language, that person speaks some language -- we still do not know which. If we instead consider a safe version of this rule, i.e., $\forall x, y\; \exists y, z_1, z_2 : \textit{birthCountry}(x, z_1) \wedge \textit{nationality}(x, z_2) \wedge \textit{officialLang}(z_2, y) \Rightarrow \textit{speaks}(x, y)$, we can now predict the person speaks the official language of their country of citizenship. If no variable is existentially quantified, i.e., each variable appears in at least two different atoms, we say the rule is \emph{closed}. Most applications relying on rules resort to safe rules, or more frequently to closed rules such as $\forall x, y, z_2\; : \textit{birthCountry}(x, z_2) \wedge \textit{nationality}(x, z_2) \wedge \textit{officialLang}(z_2, y) \Rightarrow \textit{speaks}(x, y)$.

As mentioned before, safe rules can be used to draw specific conclusions, e.g., deduce the nationality of a person from known information in a KG $\mathcal{K}$. To do so we need to introduce the notion of substitution. A \emph{substitution} $\sigma: \mathcal{V} \to \mathcal{E}$ is partial mapping from variables to constants. Applying a substitution to an atom replaces the variables of the rule by the constants associated to those variables in the substitution. We call the result of this operation an \emph{instantiation}. This operation can be naturally ported to conjunctions of atoms, which returns a set of instantiations. If we apply the substitution $\sigma = \{ x \to \mathit{E. Macron}, z_1 \to \mathit{France}, y \to \mathit{French} \}$ to the conjunction $\body: \textit{birthCountry}(x, z_1) \wedge \textit{officialLang}(z_1, y)$, we obtain instantiations that are actual facts (no variables left) $\sigma(\body) = \{ \mathit{birthCountry}(\mathit{E. Macron}, \textit{France}), \mathit{officialLang}(\mathit{France}, \mathit{French}) \}$. 
Let $R$ be a rule of the form $\hornrule$ and $\sigma$ an instantiation. We say $R$ and $\sigma$ \emph{fire} in KG $\kg$, if $\sigma(\body) \subseteq \kg$, denoted by $\sigma(\body) \Vdash \kg$. Put differently an instantiated rule fires in a KG if all the instantiated body atoms are facts from the KG. If additionally $\sigma(H) \in \kg$, we say the rule \emph{predicts} the fact obtained by instantiating $H$, which we denote by $\sigma(R) \Vdash \mathcal{K}$. That is the case for our example rule and the KG in Figure~\ref{fig:sample_kg} since both $\sigma(\textbf{B})$ and $\sigma(H) = \textit{speaks}(\textit{E. Macron}, \textit{French})$ are in the KG. 

Logical rules in KGs are unlikely to make correct predictions every time they fire. That is why we usually talk about \emph{soft rules}, that is, rules with exceptions. Take as an example the rule $\textit{birthCountry}(x, z_1) \wedge \textit{officialLang}(z_1, y) \Rightarrow \textit{speaks}(x, y)$. It is easy to see that such a rule is overall accurate but may have an exceptions, e.g., people born in a country but raised elsewhere. In a related note we should be skeptical about rules that fire or hold in very few cases. It follows from these observations that we need metrics to quantify the predictive power of rules before using them in applications. A popular score to quantify the significance of a rule $\hornrule$ in a KG $\kg$ is the \emph{support}, defined as:
\begin{equation} \label{eq:support}
    \mathit{supp}_{\kg}(\textbf{B} \Rightarrow H) = \# \sigma_H: \sigma_H(\textbf{B} \Rightarrow H) \Vdash \mathcal{K}.
\end{equation}

\noindent In this equation $\#\sigma_H$ is the number of unique instantiations $\sigma_H: \mathit{vars}(H) \to \mathcal{E}$, that is, instantiations that map the rule head variables to constants in the KG. Intuitively the support is the number of observed predictions of the rule in the KG. Those predictions are the \emph{positive examples} of the rule. The higher the support, the more evidence about the soundness of the rule we have. If we take as eample the rule 
\begin{equation}
\label{eq:exrule}
 R: \textit{birthCountry}(x, z_1) \wedge \textit{officialLang}(z_1, y) \Rightarrow \textit{speaks}(x, y),   
\end{equation}
 we can see that its support is
\[
    \mathit{supp}_{\mathcal{K}}(R) = \# (x, y): \exists z_1:  \textit{birthCountry}(x, z_1) \wedge \textit{officialLang}(z_1, y) \wedge \textit{speaks}(x, y).
\]
\noindent In our example graph of Figure~\ref{fig:sample_kg}, this value is 2 because of the substitutions $\{ x \to \mathit{E. Macron}, y \to \mathit{French} \}$ and $\{ x \to \mathit{U.v.d. Leyen}, y \to \mathit{German} \}$. 
A popular variant of the support normalizes Equation~\ref{eq:support} by the number of facts in the head relation. This is called the \emph{head coverage}:
\begin{equation} \label{eq:headCoverage}
    \textit{hc}_{\kg}(\hornrule)=\frac{\textit{supp}(\hornrule)}{\#r(x, y): r(x, y) \in \kg}.
\end{equation}

Our example rule has a head coverage of $\frac{2}{3}$ in Figure~\ref{fig:sample_kg} since it predicts 2 out of the 3 facts of the relation \textit{speaks}. The support and head coverage are anti-monotonic scores. This means that adding atoms to an existing rule cannot increase its support. This property is crucial when designing efficient algorithms that learn those rules automatically.

But even if a rule has many supporting positive examples, it may also have many counter-examples, to put it another way, instantiations for which the rule fires but its predictions are false. If those false predictions outnumber the correct predictions, then we should take the rule's predictions with a grain of salt. This illustrates the risk of learning only from positive examples. The \emph{confidence} score solves this issue by normalizing the support by the total number of examples of the rule, both positive and negative:

\begin{equation}\label{eq:conf}
    \mathit{conf}_{\mathcal{K}}(\hornrule) = \frac{\textit{supp}(\hornrule)}{\textit{supp}(\textbf{B} \Rightarrow H) + (\# \sigma_H: \sigma_H(\body) \Vdash \mathcal{K} \wedge \sigma_H(H) \nVDash \mathcal{K})}.
\end{equation}

\noindent The expression on the right-hand side of the denominator describes the number of examples that make the rule fire but whose predictions, represented by $\sigma_H(H)$, are not entailed by the KG. By ``not entailed'' we mean they contradict what is stated in the data. Bear in mind, however, that KGs do not store negative information. They cannot state things like ``Emanuel Macron does not speak English''. One could argue that the absence of such fact in our example graph (Figure~\ref{fig:sample_kg}) entails that Macron does not speak English. Unfortunately most KGs, and most particularly Web-based KGs, are inherently incomplete: the absence of evidence is not evidence of absence. This means that if the KG does not say anything about Macron speaking English, we cannot conclude he does not speak English. This information is unknown. Take as an example the rule $\mathit{worksFor}(x, \mathit{EU}) \Rightarrow \mathit{speaks}(x, \mathit{English})$. Since E. Macron works for the EU, this rule predicts he speaks English. We therefore have a problem when computing the confidence of this rule because we do not have a way to say whether this prediction is an example or a counter-example.  As we will see in the next paragraph, one needs to make some assumptions about the completeness of KGs in order to evaluate the accuracy of logical rules and their inferences.   



\paragraph{Closed World vs Open World Assumption}
The Closed World Assumption (CWA), commonly applied in database systems and logic programming, is the assumption that KGs are complete. This means that any statement not explicitly present in the KG is assumed false. This applies for $\mathit{speaks}(\textit{E. Macron}, \mathit{English})$ in our example graph from Figure~\ref{fig:sample_kg}.  It follows that this fact would be considered as a counter-example for the rule $\mathit{worksFor}(x, \mathit{EU}) \Rightarrow \mathit{speaks}(x, English)$. The CWA allow us to devise a confidence metric for rules, hence Equation~\ref{eq:conf} becomes:
\begin{equation}\label{eq:stdconf}
    \mathit{std\text{-}conf}_{\kg}(\hornrule) = \frac{\textit{supp}(\hornrule)}{\# \sigma_H: \sigma_H(\textbf{B}) \Vdash \kg }.
\end{equation}
\noindent The CWA confidence, also called \emph{standard confidence} normalizes the support of the rule by the number of examples that make the rule fire. This denominator term includes both the positive examples and the predictions that are not present in the KG -- now assumed as negative examples. In our example KG of Figure~\ref{fig:sample_kg}, the rule 
$\mathit{birthCountry} (x, \mathit{EU}) \Rightarrow \mathit{speaks}(x, English)$ has a confidence of $\frac{2}{3}$ because Angela Merkel satisfies the conditions of the rule but the conclusion is not in the KG. Albeit sensible for databases and classical association rule mining, the CWA is problematic for KGs, because they are inherently incomplete and operate under the \emph{Open World Assumption} that says nothing about absent statements: they are unknown.
\begin{figure}
    \centering
    \includegraphics[width=0.8\textwidth]{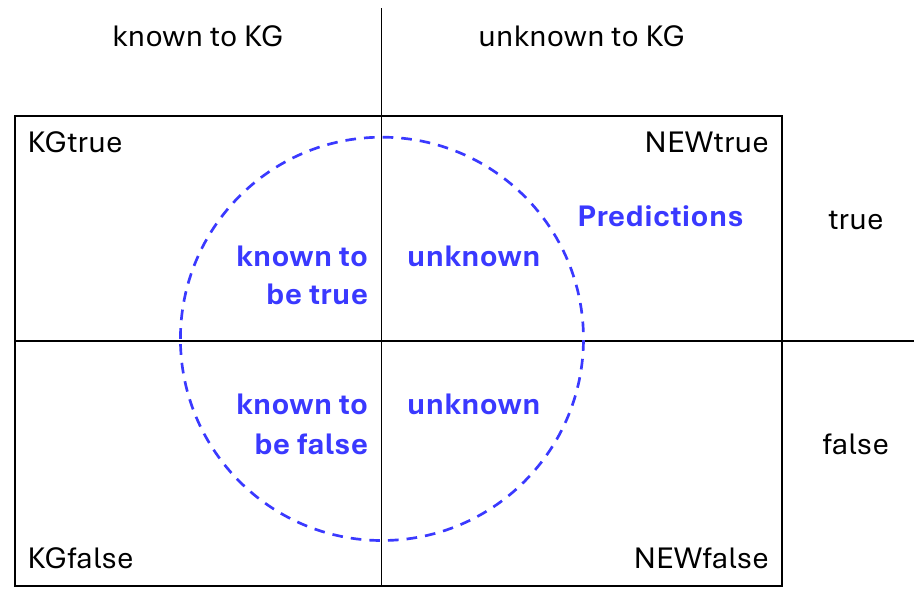}
    \caption{The classification of predictions made by a rule $B_1 \wedge B_2 \wedge ... \wedge B_n \implies p(x, y)$ pertains to a fact in the rule head. This fact can be either true or false in the real world and can be known or unknown to the KG. This results in four possible situations regarding the KG and the real world. Predictions made by the rule with respect to the KG are represented inside the circle.}
    \label{fig:kbnew}
\end{figure}

In contrast, the OWA does not offer a solution to the problem of determining whether a prediction should be counted as a counter-example or not (see Figure~\ref{fig:kbnew}). A way to deal with this issue is the \emph{partial completeness assumption}~\cite{galarraga_amie_2013}, also called the \emph{local closed world assumption}~\cite{dong2014knowledgevault}. The PCA assumes that information in KGs is added in ``batches''. If the KG constructor included one language for let us say, U.v.d. Leyen, then it included \emph{all} her languages in the KG. This assumption allows us to devise a new criterion to define counter-examples for rules: if a rule such as $\mathit{worksFor}(x, \mathit{EU}) \Rightarrow \mathit{speaks}(x, English)$ predicts a new language for a person and that language is different from the languages stated in the KG, then that prediction must be a counter-example. An important corollary of this assumption is that  if the KG does not know any language for that person, then the OWA applies and that example is labeled unknown and therefore excluded both as positive example or as counter-example. This assumption can be therefore used to devise a new confidence score, the \emph{PCA confidence}~\cite{galarraga_amie_2013}:
\begin{equation}\label{eq:pca}
    \mathit{pca\text{-}conf}_{\kg}(\hornrule) = \frac{\textit{supp}(\hornrule)}{\# \sigma_H: \sigma_H(\body \wedge H') \Vdash \kg },\;\;H'=r(x, y')\;\;\text{or}\;\;H'=r(x', y).
\end{equation}
Equation~\ref{eq:pca} normalizes the support by the number of head substitutions that make the rule fire and for which there is a known head fact. This fact can be the rule's prediction but can be a different since the new variables $x', y'$ in the formula are existentially quantified. In our example from Figure~\ref{fig:sample_kg}, the PCA confidence of the rule $\mathit{speaks}(\textit{E. Macron}, \mathit{English})$ is $\frac{2}{2}$. Differently from the standard confidence, Angela Merkel is not used as a counter-example because the KG does not know anything about the languages she speaks, i.e., there is no match for the head atom. Furthermore, the PCA confidence assumes the prediction is made in one direction, usually the more functional direction: we predict the language of a person instead of predicting all the speakers of a language. Therefore the choice of the atom $H'$ depends on the nature of the head relation.

As shown by Galárraga et al~\cite{galarraga_fast_2015}, the PCA is a sensible assumption for many relations, specially functional (1-1) or quasi-functional (almost 1-1) relations such as place of birth or nationality. The PCA is clearly less sound for n-to-n relations such as friendships or co-authorships and can make mistakes for quasi-functions. In our example graph (Figure~\ref{fig:sample_kg}) the PCA is right about the nationalities, birth countries and genders of our example entities, but it is false for the languages they speak since it would assume E. Macron does not speak English and U.v.d. Leyen does not speak any other languages besides English and German -- which is actually not true. 

\section{Rule Mining}
\emph{Rule mining} is the task of learning logical rules from a knowledge graph fully automatically. This problem is challenging for two main reasons. First, it is computationally expensive, specially for large KGs, because it incurs the exploration of a very large search space. Second, it requires to make assumptions about what constitutes a counter-example in order to evaluate the quality of the rules. Since those rules are usually used for inference tasks, it is common to focus on closed rules, which as a side effect also reduce the space of rules to explore. There are different approaches to rule mining that we will describe in the following.


\subsection{Inductive Logic Programming}
The study of learning Horn clauses has been a significant focus in the inductive logic programming (ILP) field~\cite{muggleton_inductive_1994,DBLP:books/sp/NienhuysW97,raedt_inductive_2017}.
ILP methods are based on search of the space of possible patterns or rules. 
To systematically explore the space during the learning process, ILP methods usually use \emph{refinement operators}.  
A refinement operator defines a way to move from one candidate rule to another, more specific or more general. 
In rule mining, particularly within ILP, refinement operators help navigate the pattern or rule space in a structured manner.
Refinement operators usually add/remove atoms or specialize/generalize predicates. We will see example refinement operator in Section~\ref{sec:amie} where the algorithm AMIE is discussed.   

\subsubsection{AMIE}
\label{sec:amie}
 AMIE \cite{galarraga_amie_2013} is a top-down closed rule mining algorithm designed for large KGs under the OWA. AMIE constructs new rules by adding atoms to already discovered rules. This process is called refinement. Some of those rules will be intermediate non-closed rules that AMIE refines but does not output. By default, the algorithm starts with all the rules of the form $\emptyset \Rightarrow r(x, y)$ that are iteratively refined via three mining operators:
 \begin{description}
     \item[Add dangling atom] refines a rule with a new atom containing a fresh variable. For rule $\emptyset \Rightarrow speaks(x, y)$, this operator could produce rules such as $\mathit{hasOfficialLang}(z_1, y) \Rightarrow \mathit{speaks}(x, y)$ or $\mathit{nationality}(x, z_1) \Rightarrow \mathit{speaks}(x, y)$.
     \item[Add closing atom] adds a new atom without fresh variables. In our previous example, this operator could lead to the rule $\mathit{likes}(x, y) \Rightarrow \mathit{speaks}(x, y)$. Similarly, the rule $\mathit{hasOfficialLang}(z_1, y) \Rightarrow \mathit{speaks}(x, y)$ could be closed by adding the atom $\mathit{nationality}(x, z_1)$ to the body.
     \item[Add instantiated atom] refines the rule with an instantiated atom, that is, an atom where one of the arguments is a constant, e.g., the rule $\mathit{likes}(x, y) \Rightarrow \mathit{speaks}(x, y)$ could become $is(x, Linguist) \wedge \mathit{likes}(x, y) \Rightarrow \mathit{speaks}(x, y)$. By default, this operator is disabled, but can be turned on by the user. This also allows the system to start with rules of the form $\emptyset \Rightarrow r(x, C)$ or $\emptyset \Rightarrow r(C, y)$ for constants $C$ from the KG.
 \end{description}
 These mining operators allow AMIE to explore the space of closed Horn rules. AMIE imposes a user-defined minimum threshold on support and a maximum rule length (also configurable by the user) to keep the search space under control. By default, AMIE finds rules up to 3 atoms and stops the refinement as soon as head coverage drops below 1\%. This policy, based on the anti-monotonicity of the head coverage, speeds up the mining by avoiding noisy rules that cover too few positive examples. AMIE can also enforce user-defined thresholds on standard and PCA confidence (set by default to 10\%). 


All these considerations made AMIE the fastest rule mining algorithm on KGs at its time of publication, achieving a speed up of at least 3 orders of magnitude w.r.t. classical inductive logic programming approaches such as WARMR~\cite{goethals2002warmer} and ALEPH~\cite{muggleton1996aleph} on modern KGs such as YAGO2~\cite{yago} (approx. 1M facts) -- previous approaches could not handle larger datasets such as DBpedia. Moreover, and in contrast to its competitors, its reliance on the PCA confidence to quantify the quality of rules made it produce more and more precise predictions than its competitors. On YAGO2, for instance, the rules had a precision in the range of 30\%-40\% and the PCA confidence proved more suitable than the standard confidence at ranking best the rules that inferred good new predictions -- predictions beyond the KG.

\paragraph{AMIE+~\cite{galarraga_fast_2015}} 
improves over AMIE with the help of various algorithmic optimizations to speed up rule mining. This included changes in the rule refinement procedure as well as some heuristics to discard potentially noisy rules. AMIE+ avoids refining rules when the resulting refinement cannot lead to a closed rule given the maximal length constraint. It also implements some query simplification for recursive rules, i.e., rules where a predicate appears more than once. It also implements a skyline technique that stops refining closed rules that have already attained 100\% confidence and propose lower bounds and confidence estimations that prune noisy rules, i.e., rules of very low confidence, before computing their actual confidence scores -- a computationally expensive task. All these optimizations allowed AMIE+ to run on larger datasets such as DBpedia 3.8 and Wikidata 2014 and achieve a speed up of at least one order of magnitude w.r.t. AMIE. 


\paragraph{AMIE 3}
Lajus et al.~\cite{lajus2020fast} introduced the latest version of AMIE, called as AMIE3, that features several query processing and data representation improvements. For example, the \emph{existential variable detection} heuristic optimizes the queries required to compute the rule confidence scores by properly identifying variables with existential semantics whose instantiations do not need to be fully enumerated. AMIE3 also proposes a lazy evaluation criterion for the confidence scores. This strategy stops the enumeration of the solutions of the normalization term (denominator) of the confidence scores as soon as it is clear that the resulting confidence will be below the minimum confidence threshold. Other optimizations include the parallelization of the construction of some indexes and the use of an integer-based representation for the entities and triples of the KG.     
AMIE3 is then further compared with modern rule miners such as Rudik~\cite{ortona_robust_2018} and ScaleKB~\cite{yang2016scalekb} on YAGO2, DBPedia, and Wikidata and find that AMIE3 is one order of magnitude faster. 


\subsection{Path Sampling and Generalization}
\paragraph{RuDiK~\cite{ortona_robust_2018}} is an algorithm that discovers both positive and negative rules. Mining negative rules helps to detect erroneous facts, which can be common in knowledge bases, due to errors being propagated. RuDiK consists of three modules: the first module generates negative examples, the second one is an incremental rule miner and the last module executes rules, to generate new facts and find inconsistencies.

\begin{description}
    \item[Negative example generator] creates negative examples, given a knowledge base and a target predicate. It does this by leveraging the Local Closed-World Assumption (LCWA) \cite{dong2015data}. Under this assumption, if a triple $q(s,o)$ does not occur in a knowledge base, but $q(s,x)$ is present, then $q(s,o)$ is false. Similarly, if $q(s,o)$ is present but $q'(s,o)$ is not, then $q'(s,o)$ is false. RuDiK finds entities whose information is more likely to be complete, to generate good negative examples.
    \item[Incremental rule miner]  discovers Horn Rules.
    Here, the atoms are of the form $q(s,o)$. The algorithm uses a set of positive examples $G$ and a set of negative examples $V$. The ideal solution is the minimal set of rules for which all the examples in G are valid and none of those in $V$ are. 
    The goal of this rule miner is to find the optimal set of weighted rules. The weight of a rule has two components. The first is the ratio between the coverage of this rule over $G$ and $G$ itself. The second component measures the same thing over $V$. Parameter $\alpha$ is set to define the weight of the first component. There is also $\beta$, defined as 1-$\alpha$. $\beta$ defines the importance of the second component. This definition of weight is extended to define a marginal weight as follows. If $R$ is a set of rules and $r$ is a rule :
    \begin{equation}
        w_{m}(r) = w(R \cup r) - w(R)
    \end{equation}
    
    A rule will not be added to the solution if its marginal weight is at or below 0. A valid rule r(x,y) can be represented as a path between x and y in the knowledge base, which is represented as a directed graph. Another type of atoms can be included in the rules : literal comparison. For example, the rule\\ $bornIn(a, x)$ $\land$ x $\neq$ U.S.A. $\implies$ $\neg$ $president(a, U.S.A.)$.\\
    
    The algorithm starts with an entity x and keeps a set of candidate paths. At each step, the path with the smallest marginal weight is expended. Once a path is considered valid, it is added to the solution, and it stops being expended.
    \item[Rule execution] Once a rule has been discovered, RuDiK can run it in the knowledge base, as a SPARQL query. This enables it to deduce new facts, or to detect erroneous ones that are in the knowledge base. The accuracy for new facts is 85\%, and 97\% for inconsistencies. When running RuDiK over Wikidata, DBpedia and YAGO 3, the proportion of erroneous triples was respectively 0.23\%, 0.26\% and 0.6\%.
    
\end{description}


Different parameters impact on the performance of RuDiK. 
Turning off literal comparison visibly degrades accuracy for both positive and negative rules.  The level of noise in the database also affects the performance, though RuDiK is rather robust in this regard. Another important parameter is the maximum path length. When it is set at 2, precision for positive rules drops to 49\%. When it is set at 4, RuDiK does not finish after 24 hours, and the precision is not notably better than when it is set at 3. Therefore, 3 is used as the maximum path length. The last parameter is $\alpha$ and $\beta$, the weight parameters. There are different optimal values for positive and negative rules, however the algorithm is robust and the variation in performance are limited as long as $\alpha$ (and $\beta$) is in the [0.1,0.9] range.

\paragraph{AnyBURL~\cite{meilicke2019introduction}}
is a bottom-up algorithm inspired by Golem \cite{muggleton1990efficient} and Aleph \cite{muggleton1996aleph}. The algorithm chooses random paths of a set path profile, and generalizes them to rules. A path profile ``describes path length and whether the path is cyclic or acyclic". The support and confidence of these rules are then computed, and the rules that fulfil a given criteria, usually minimum support or confidence, are stored. Given a completion task, the candidates are ranked by the maximum confidence of the rules that have generated them, then by the second best one, and so on, until one rule stands out. There is more than one round of mining and selection of rules. Each round lasts a set amount of time. At the end, the length of possible paths can be increased if the set of results reaches {\it saturation}, i.e. if the number of new rules being discovered is too low.

AnyBURL also uses reinforcement learning to determine how much effort should be dedicated for each path profile. It also uses {\it Object Identity}, which assumes that when two variables appear in a rule, they designate different entities. This is made to avoid redundant rules. There is a new version of AnyBURL \cite{meilicke_anytime_2024}, which introduces four major modifications. The first modification is object identity as described in \cite{semeraro1994avoiding}, to avoid learning redundant rules, which would skew the confidence score. The second modification is confidence sampling, which serves to avoid a depth-first search when computing the confidence of a rule in a very large dataset. The third modification is reinforcement learning-based sampling which makes the path sampling more robust, and especially less sensitive to change in parameter settings. The fourth modification is multi-threading which enhances the performance of AnyBURL on very large graphs and makes it 20 times faster than SOTA.

\paragraph{RARL~\cite{pirro_relatedness_2020}} uses rule relatedness (or TBox relatedness) to rank candidate rules. The authors define two boxes. The ABox contains the facts, while the TBox contains the underlying schema of the knowledge base. Rules are considered related when they often link the same subjects and objects in the ABox. The confidence of these rules is computed under the Partial Completeness Assumption \cite{galarraga_fast_2015}, which allows the creation of negative examples.

\subsection{Linear programming}
A recent paper \cite{dash2023rule} presents LPRules, an algorithm for rule mining that uses linear programming to mine rules inside a knowledge graph. It creates a weighted linear combination of FOL rules that are then used as a scoring function for knowledge graph completion. It also limits the size of the rules, to increase human interpretability.

\section{Neurosymbolic Methods} 

\subsection{Deep learning and Rules} 

We now introduce end-to-end approaches that utilize deep neural networks (DNNs) to learn rules by optimizing objective functions approximating path patterns.

\subsubsection{Neural Logic Programming} 
\label{ss:nlp}
\emph{Neural LP}~\cite{yang_differentiable_2017} is one of the pioneering efforts to integrate rule structure learning with parameter learning in an end-to-end differentiable model. 
It draws inspiration from the \emph{TensorLog}~\cite{DBLP:journals/corr/Cohen16b} differentiable probabilistic logic framework. 
The TensorLog framework compiles rule inference into a series of differentiable operations by linking rule application to sparse matrix multiplications.
The method thus simplifies the rule learning problem to algebraic operations on neural embedding-based representations of a given knowledge graph.

The reasoning task the NeuralLP addresses involves three components: \emph{query}, an entity \emph{tail} about which the query is made, and an entity \emph{head} that serves as the query's answer. The objective is to generate a ranked list of entities in response to the query, aiming to position the correct answer (i.e., the head) as high as possible on this list.
We can formalize the query as a rule 
\begin{equation}
q_1(x, z_1) \wedge ... \wedge q_n(x, z_n) \Rightarrow p(x, y)  
\end{equation}
with associated confidence $\alpha \in [0, 1]$, where $p(x, y)$ is the query, and $q_1, ..., q_n$ are relations in the knowledge base. 
During inference, given an entity $x$, the score for each entity $y$ is calculated as the sum of the confidence scores of rules that imply $p(x, y)$. A ranked list of entities is then returned, where a higher score corresponds to a higher ranking. 

\paragraph{TensorLog}
\label{lbl:TensorLog}
TensorLog maps each entity $e_i \in \mathcal{E}$ to a one-hot vector $\mathbf{v}_i \in \{0, 1\}^{|\mathcal{E}|}$ where only the $i$-th entry is 1, and it defines an operator $M_q$ for each relation $q$ by mapping each relation $q \in \mathcal{R}$  to a matrix $\mathbf{M}_q \in \{0,1\}^{|\mathcal{E}|\times|\mathcal{E}|}$ such that its $(i,j)$ entry is 1 iff $q(e_i,e_j)$ is a fact in the KG, where $e_i, e_j \in \mathcal{E}$. So $\mathbf{M}_q$ is essentially an adjacency matrix.

For instance, considering a subgraph of the $KG$ presented in Figure~\ref{fig:sample_kg} consisting of 5 entities, every  entity  is encoded as a one-hot vector of length 5, corresponding to the number of the entities in the subgraph, so, for relations $q_1 = \textrm{birthCountry}$, $q_2 = \textrm{officialLang}$ 
we have the following adjacency matrices: 

\[
\mathbf{M}_{q_1}  = 
\begin{blockarray}{ccccccc}
 & \textrm{E. Macron} &  \textrm{U.v. Leyen} & \textrm{EU} & \textrm{French} & \textrm{France} \\
\begin{block}{c[ccccc]c}
 & 0 & 0 & 0 & 0 & 1\bigstrut[t] & \textrm{\textbf{E. Macron}} \\
 & 0 & 0 & 0 & 0 & 0 &            \textrm{U.v. Leyen} \\
 & 0 & 0 & 0 & 0 & 0 \bigstrut[b] &  \textrm{EU}\\
 & 0 & 0 & 0 & 0 & 0\bigstrut[b] & \textrm{French}\\
 & 0 & 0 & 0 & 0 & 0\bigstrut[b] & \textrm{France}\\
\end{block}
\end{blockarray}\vspace*{-1.25\baselineskip}
\]

\[
\mathbf{M}_{q_2}  = 
\begin{blockarray}{ccccccc}
 & \textrm{E. Macron} &  \textrm{U.v. Leyen} & \textrm{EU} & \textrm{French} & \textrm{France} \\
\begin{block}{c[ccccc]c}
 & 0 & 0 & 0 & 0 & 0\bigstrut[t] & \textrm{E. Macron} \\
 & 0 & 0 & 0 & 0 & 0 &            \textrm{U.v. Leyen} \\
 & 0 & 0 & 0 & 0 & 0 \bigstrut[b] &  \textrm{EU}\\
 & 0 & 0 & 0 & 0 & 0\bigstrut[b] & \textrm{French}\\
 & 0 & 0 & 0 & 1 & 0\bigstrut[b] & \textrm{France}\\
\end{block}
\end{blockarray}\vspace*{-1.25\baselineskip}
\]


We now establish the connection between TensorLog operations and logical rule inference, where the goal is to imitate logical rule inference for some entity $e_i$.
The application of the rule on an entity $e_i$ can be represented by performing matrix multiplications 
\begin{equation}
\mathbf{M}_{q_1} \cdot \mathbf{M}_{q_2} \cdot ... \cdot \mathbf{M}_{q_n} \cdot \mathbf{v}_i = \mathbf{s}  \end{equation}
For example, consider the rule 
\begin{equation}
birthCountry(x,z), officialLang(z,y) \Rightarrow speaks(x, y)
\end{equation}  
which we can translate, for the sake of inference, to: 

\begin{equation} \mathbf{M}_{birthCountry}\  \mathbf{M}_{officialLang}\ \mathbf{v}_y = \mathbf{s} 
\end{equation}
The non-zero entries in the  vector $\mathbf{s}$ point to the entities for which $p(x,y)$ (in this case $speaks(x, y)$) is derived.  
These non-zero entries of the vector $\mathbf{s}$ equals the set of $y$ such that there exists $z$ that $birthCountry(x, z)$ and $officialLang(z, y)$ are in the $KG$.

\[
\mathbf{M}_{q_1 \times q_2}  = 
\begin{blockarray}{ccccccc}
 & \textrm{E. Macron} &  \textrm{U.v. Leyen} & \textrm{EU} & \textrm{\textbf{French}} & \textrm{France} \\
\begin{block}{c[ccccc]c}
 & 0 & 0 & 0 & 1 & 0\bigstrut[t] & \textrm{E. Macron} \\
 & 0 & 0 & 0 & 0 & 0 &            \textrm{U.v. Leyen} \\
 & 0 & 0 & 0 & 0 & 0 \bigstrut[b] &  \textrm{EU}\\
 & 0 & 0 & 0 & 0 & 0\bigstrut[b] & \textrm{French}\\
 & 0 & 0 & 0 & 0 & 0\bigstrut[b] & \textrm{France}\\
\end{block}
\end{blockarray}\vspace*{-1.25\baselineskip}
\]

By assigning $\mathbf{v}_x = [1, 0, 0, 0, 0]^{\top}$ to point to $\textrm{E. Macron}$ and performing the matrix multiplications, as the result we have $\mathbf{s} = [0, 0, 0, 1, 0]^{\top}$, which points to $\textrm{French}$.




Let $\beta_i$ denote an ordered list of all relations appearing in the rules. 
Following~\cite{yang_differentiable_2017}, the inference for each query is defined, more generally, as:
\begin{equation}
\label{eq:queryinference}
    \sum_l\alpha_{l}\prod_{k \in \beta_l}\mathbf{M_{q_k}}
\end{equation}

\begin{equation}
\label{eq:inference}
\mathbf{s}=\sum_l(\alpha_{l}(\prod_{k \in \beta_l}\mathbf{M_{q_k}}\mathbf{v_y})),\ \ score(x|y) = \mathbf{v_x}^{\top}\mathbf{s} 
\end{equation}

In summary, the learning problem for each query becomes: 
\begin{equation}
 \max_{\{\alpha_l, \beta_l\}} { \sum_{\{x,y\}} score(x|y) = \max_{\{\alpha_l, \beta_l\}}  \sum_{\{x,y\}}\mathbf{v}_x^{\top} \left(\sum_l (\alpha_l ( \prod_{k \in \beta_l} \mathbf{M}_{q_k} \mathbf{v}_y )) \right) }    
\end{equation}
  
The goal is to extract the rules from the solution of the above optimization problem, by using the defined operators. 

\paragraph{Learning rules}

The set of rules that imply each query and the confidences associated with these rules need to be learnt, that is $\{\alpha_l, \beta_l\}$ are to be learnt. 
To facilitate this by addressing the problem of enumerating rules, Yang et al.~\cite{yang_differentiable_2017} proposes to rewrite Equation~\ref{eq:queryinference} in the following way:
\begin{equation}    \prod_{t=1}^{T}\sum_k^{|\mathcal{R}|}\alpha_t^k\mathbf{M}_{q_k}
\end{equation}
where $T$ denotes the maximum length of rules and $|\mathcal{R}|$ the number of relations in the knowledge graph. 
In order to combine the enumeration of the rules and confidence assignment, the key difference in this new formulation is that each relation in the rules is associated with a weight. 

Since rules may be of different length Yang et al~\cite{yang_differentiable_2017} introduced a recurrent formulation that resembles this in Equation~\ref{eq:inference}. 
This version uses auxiliary memory vectors $\mathbf{u}_t$, which are at the beginning set to the given entity $\mathbf{v}_y$:
\begin{equation}
    \mathbf{u_0} = v_y
\end{equation}
Then, at each step, the model first computes a weighted average of the previous memory vectors using the memory attention vector $\mathbf{b}_t$, and secondly, it applies the TensorLog operators using the operator attention vector $\mathbf{a}_t$:
\begin{equation}
    \mathbf{u_t} = \sum_k^{|\mathcal{R}|}a_t^k\mathbf{M}_{q_k} \left( \sum_{\tau=0}^{t-1}b_t^\tau\mathbf{u}_{\tau} \right)    \textrm{      for  } 1 \leq t \leq T
\end{equation}

In the last step, the model computes a weighted average of the memory vectors. In order to choose a proper rule length, attention is used in this step: 
\begin{equation}
    \mathbf{u_{T+1}} = \sum_{\tau=0}^{T}b_{T+1}^\tau\mathbf{u}_{\tau}
\end{equation}

The learnable parameters are the memory and operator attention vectors. Recurrent neural networks can now be used that fit this recurrent formulation, and the authors of ~\cite{yang_differentiable_2017} used LSTM for this purpose.

\emph{Neural-Num-LP}~\cite{wang_differentiable_2020} enhances Neural-LP by incorporating the ability to learn rules with negations and numeric values. Additionally, it improves on Neural-LP through implicit representation of essential matrix operations. These improvements include the use of dynamic programming, cumulative sums for numerical comparison features, and low-rank factorisations for negated atoms. 

\emph{DRUM}~\cite{sadeghian_drum_2019} was introduced to mitigate a tendency of Neural-LP to learn meaningless rules with high confidence that share atoms with valid rules. To mitigate this issue, DRUM employs bidirectional RNNs to prune potentially incorrect rules as well as low-rank decompositions of matrix $M_p$.  

\subsubsection{Decoupling Models}
To address challenges of optimization in joint learning of rule structures and confidence, several methods have been proposed that separate these two tasks. 

\emph{RNNLogic}~\cite{qu_rnnlogic_2020} addresses the challenges in existing methods that struggle with  navigating a large search space (as in neural logic programming). 
RNNLogic is composed of a rule generator   and a reasoning predictor. The rule generator is responsible for structure learning and a reasoning predictor for confidence learning.
The rule generator produces logic rules for the reasoning predictor, for a given query.
The reasoning predictor uses the generated rules as input to reason over a knowledge graph and predict the answer. In each iteration, the rule generator produces a set of logic rules. Moreover, in each iteration, the reasoning predictor is updated to explore these rules for reasoning.
In the next step, a set of high-quality rules is identified from the generated rules via posterior inference. 
In the final step, the rule generator is updated to align with the high-quality rules identified in the previous step. 

RNNLogic is optimized using an Expectation-Maximization (EM) algorithm.

\emph{RLogic}~\cite{cheng_rlogic_2022} is a method for mining chain-like rules.   
The authors highlight two limitations of other algorithms: their  dependence on observed rule instances to define the score function for rule evaluation, and their inability to mine rules that lack support from rule instances. 
To address these challenges, RLogic operates by sampling closed paths within a knowledge graph and proposes a sequential rule learning algorithm that decomposes a sequential model into smaller atomic models in a recursive manner.
For example, the relation path $[birthCountry, officialLang]$ existing in our sample KG (Figure~\ref{fig:sample_kg}) can be replaced by a single relation $nativeLang$. To address cases when  the relation to replace with  might not be present in the knowledge graph, a ''null'' predicate is also introduced into the relations set. 

The authors introduce a \emph{relation path encoder} and a \emph{close ratio predictor}.
The goal of the relation path encoder is to find a head relation $p_h$ to replace an entire relation path.
The relation path encoder reduces the rule body $[q_1, . . . , q_n]$  to a head $p_h$ by recursively merging relation pairs using a greedy algorithm.
The close ratio predictor is based on the observation that, even after logically deducing a reduction of the relation path to a single relation head, this head relation may not always be present in the knowledge graph. Therefore, the task of close ratio predictor is to estimate the ratio that a path will close and the probability of replacing a relation pair with a single relation. A two-layer, fully connected neural network (MLP) is used for this purpose. 

\subsection{Embeddings and rule learning} 
\label{sec:embeddings-rules}
Among other things, the previously mentioned algorithms have been designed to provide a rule-based approach for solving predictive tasks, e.g. the KG completion, with a set of mined rules. The main feature of rule-based systems is the need to first obtain rules whose relevance is then computed based on the coverage of a given rule by some examples occurring in the input KG. Hence, the rules searching process and their relevance determination often require storing the entire KG in the memory to allow for fast exploration of the search space or walking through the graph. This may be a problem for large KGs since they have high resource requirements, and the existing systems are not able to effectively scale input data and the mining process.

Graph embeddings are often regarded as an alternative to rule-based approaches for solving specific prediction tasks over graph data, e.g. for link prediction. The graph-embeddings prediction model is composed of a nodes/relations representation (e.g. vectors, matrices) and a scoring function (to calculate the reliability of a predicted entity). The popularity of these kinds of models is given by a simple vector or matrix representation of the entire graph where fast and scalable vector operations can be performed, e.g., to determine similarities among nodes within Euclidean space. Recent studies have also shown that some techniques using graph embeddings outperform convenient rule-based approaches, like AnyBURL \cite{rossi2021knowledge, meilicke2018fine}. Some well-known methods to transform a KG or its individual components (nodes and edges) into vectors are, e.g. RESCAL \cite{nickel2011three}, HolE \cite{nickel2016holographic} and TransE \cite{bordes2013translating}. Besides pure graph embedding models, some algorithms even combine the rule-based with the graph embedding approach.

An early approach combining embeddings and rule-based systems was called EmbedRULES \cite{embedRules}.   
The RLvLR algorithm \cite{Omran:2018:SRL:3304889.3304958} uses low-dimension embeddings of RDF KG resources and predicates for fast search of Horn rules. This algorithm, which according to the authors' benchmark, outperforms EmbedRULES, focuses on a specific predicate $p$ at the head position. For each $p$ it creates a sample of an input KG with such facts that are connected to $p$ up to the maximum length of the rule. This operation is required for a large KG since RLvLR uses the RESCAL factorization to create embeddings by default, which can be slow for large data sets. Most of the mining sub-processes, such as paths finding, support and confidence computations, are performed by matrix operations from embeddings and adjacency matrices. Although this method can be faster than state-of-the-art approaches, such as AMIE, it is limited only to learning rules for a specific predicate and is not designed to discover rules with constants. The main use case of this algorithm is traceable KG completion with a given predicate.

Another rule mining system using embeddings is RuLES\footnote{https://github.com/hovinhthinh/RuLES} \cite{ho2018rule}. It uses the AMIE approach to generate rules (with or without constants) and an embedding pre-trained model by TransE, HolE, or SSP \cite{xiao2017ssp} for computing measures of significance.

While learning rules from embeddings has certain advantages, it is also known to have multiple weaknesses. 
\paragraph{Differences in predictive performance between rule-based, rule embedding and pure embedding models} There is a paucity of research showing better performance of embeddings-based rule approaches over pure rule learning approaches. Compared with the state-of-the-art RLvLR algorithm, the pure rule-learning approaches AnyBURL and its enhanced version SAFRAN \cite{ott2021safran} are reported to perform better \cite{ott2021safran}. However, this benchmark is based only on one dataset (FB15K-237). The same paper \cite{ott2021safran} also shows that SAFRAN generally performs on part with the best embedding-based (latent) approaches, but unlike them, it is rule-based and thus inherently interpretable. It should be noted the evaluation in \cite{ott2021safran}  is limited by possibly different evaluation conditions between RLvLR and SAFRAN and may not be free of bias, as the evaluation was done by the author of some of the compared methods.

\paragraph{Explainability}
With the growing emphasis on explainability in machine learning, a major limitation is that the predictions generated by graph embedding models are not traceable. Thus, the reliability of the prediction is given by the scoring function, which, however, does not explain to us what parameters lead to a given score. 
This is in contrast to rule-based models, where a specific score (confidence) value can be traced back to individual paths in the training data. For example, the RDFRules system offers a graphical interface to easily trace predictions based on AMIE-like rule-based models \cite{zeman2021rdfrules}.

\subsection{RLvLR: Rule Learning via Learning Representations}
The Rule Learning via Learning Representations (RLvLR) algorithm is inspired by NeuralLP. It mines \emph{closed} rules introduced in section \ref{sec:rules}, that have the form shown in Eq.~\ref{eq:1}. 

\begin{equation}
\label{eq:rlvlrrule}
p_1(x, z_1) \wedge p_2(z_1,z_2) ... \wedge p_n(z_n-1, y) \Rightarrow p(x, y).  
\end{equation}

RLvLR uses the  Standard confidence (Eq.~\ref{eq:stdconf}) and Head Coverage (Eq.~\ref{eq:headCoverage}) to evaluate the quality of rules. 
RLvLR authors state  three main improvements compared to previous approaches such as NeuralLP described earlier:
\begin{itemize}
    \item removing data not relevant for computation,
    \item argument embeddings: new rule quality measure through,  
    \item rule quality computed through matrix operations.
\end{itemize}

\paragraph{Removing data not relevant for computation}
This operation takes advantage of the problem formulation, where for a given head predicate $p$ and maximum rule length $l$, only entities that 
are directly or indirectly related to $p$ are relevant for the mining.

For each head predicate $p$, this procedure creates a subset of the input KG containing facts that are connected to $p$ up to the maximum length of the rule $l (l \geq 2).$
The fact and entity selection is done so that the subset contains all information relevant for learning rules of length $l$ with a given head predicate $p$. The algorithm first identifies the sets $e_0  \ldots  e_i \ldots e_{l-2}$, which contain entities in facts directly (for $i=0$) or indirectly (for $i>0$) related to $p$. Consequently a subset of the original KG is generated, referred to as $KG'$. $KG'$ contains only those facts from the original KG, where both entities in the fact (subject and object) are present among the previously identified entities (those in $E'=\bigcup_i^{l-2}e_i$).

Referring to the sample knowledge graph in Figure~\ref{fig:sample_kg}, consider this procedure for \emph{speaks} as the head predicate $p$ and $l=2$. The algorithm first identifies the set $e_0$, which contains entities in facts directly related to $p$. In this case, the directly related facts are $\{speaks(Uvd Leyen, English), speaks(Macron, French)\}$, hence  the set $e_0=\{Macron,French,Uvd Leyen, English\}$. Since $l=2$, the set $e_{l=2}'=e_0$ and $KG'_2={speaks(E Macron, French), speaks(U.v.d. Leyen, English)}$. However, there is no non-trivial rule of length 2 that can be extracted from $KG'_{l=2}$. We need to, therefore, increase the value of $l$ to $l=3$. Now, we additionally need to compute the set $e_1$, which will contain those entities that are linked to any of the entities in $e_0$ by any predicate. 
We get $e_1=\{English,German,Germany,EU,female,France,male\}$. Consequently, $e_{l=3}$ will contain all entities from the original KG except \emph{A. Merkel}. Based on $e_{l=3}$, we will get $KG'_{l=3}$, which will contain all statements in the original KG in Figure~\ref{fig:sample_kg} except for \emph{nationality(A. Merkel, Germany)} and \emph{birthCountry(A. Merkel, Germany)}. From this KG, the algorithm can extract rules such as the one in Eq.~\ref{eq:exrule} and the absence of some facts (in this case, two facts with A Merkel), will make this process faster.

Note that while this step was originally called `sampling' by RLvLR authors, it does not, in our opinion, correspond to the probabilistic implementation of sampling, which is typically understood as a random process affecting a user-set percentage of data samples. As the example shows, in RlvLR, this step can be variously effective based on the rule's length ($l$ parameter) and the graph's overall characteristics.   

\paragraph{Argument embedding}
KG embeddings, as implemented by, e.g., the RESCAL algorithm, apply to entities and relations. With \emph{argument embeddings}

RLvLR uses both synonymy and the newly introduced co-occurrence scoring functions. We will first introduce the synonymy scoring function, including with an example, and then we will briefly cover the more complex co-occurrence scoring function, details of which can be found in the article describing RLvLR \cite{omran2019embedding}.

Using the notation we introduced in section~\ref{ss:nlp},  predicates in a body of a RLvLR rule are represented using adjacency matrices $\mathbf{P}_1, \ldots, \mathbf{P}_n$ and the head predicate using embedding matrix $\mathbf{P}$, the product $\mathbf{P}_1 \cdot \mathbf{P}_2 \cdot \ldots \cdot \mathbf{P}_n$ should, according to the authors, capture pairs of entities connected by the body of the rule. As the body should be predictive of the head, the more this product is similar to the matrix $\mathbf{P}$, the better. To measure matrix similarity,  the RLvLR algorithm uses the exponentiated result of a  Frobenius norm of the difference between the two matrices. Unfortunately, the authors do not provide source code or additional details on how the embeddings are computed, besides a generic reference to predicate embeddings from \cite{yang2015embedding}.

In addition to the synonymy scoring function, the RLvLR algorithm uses the \emph{co-occurrence scoring function}. To compute this, the authors introduced the concept of argument embeddings, which are averages of the ``embedding vectors'' in the subject and object positions of the given predicate.

\paragraph{Matrix computation of rule evaluation measures}
The RLvLR algorithm uses adjacency matrices to compute the rules' standard confidence and head coverage.
We will illustrate how head coverage is computed.
  
Let us use the sample knowledge graph in Figure~\ref{fig:sample_kg}, and rule~\ref{eq:exrule} as an example. 

In the following, there are adjacency matrices for the two  predicates in the body of the rule.

\[
\mathbf{M}_{birthCountry} = 
\begin{blockarray}{lccccccccccc}
 & \textrm{UvL} & \textrm{AM} & \textrm{EM} & \textrm{Gy} & \textrm{Fr} & \textrm{EU} & \textrm{En} & \textrm{Ge} & \textrm{Fr} & \textrm{ma} & \textrm{fe} \\
\begin{block}{l[ccccccccccc]}
\textrm{U.v.d. Leyen} & 0 & 0 & 0 & 1 & 0 & 0 & 0 & 0 & 0 & 0 & 0 \\
\textrm{A. Merkel}    & 0 & 0 & 0 & 1 & 0 & 0 & 0 & 0 & 0 & 0 & 0 \\
\textrm{E. Macron}    & 0 & 0 & 0 & 0 & 1 & 0 & 0 & 0 & 0 & 0 & 0 \\
\textrm{Germany}      & 0 & 0 & 0 & 0 & 0 & 0 & 0 & 0 & 0 & 0 & 0 \\
\textrm{France}       & 0 & 0 & 0 & 0 & 0 & 0 & 0 & 0 & 0 & 0 & 0 \\
\textrm{EU}           & 0 & 0 & 0 & 0 & 0 & 0 & 0 & 0 & 0 & 0 & 0 \\
\textrm{English}      & 0 & 0 & 0 & 0 & 0 & 0 & 0 & 0 & 0 & 0 & 0 \\
\textrm{German}       & 0 & 0 & 0 & 0 & 0 & 0 & 0 & 0 & 0 & 0 & 0 \\
\textrm{French}       & 0 & 0 & 0 & 0 & 0 & 0 & 0 & 0 & 0 & 0 & 0 \\
\textrm{male}         & 0 & 0 & 0 & 0 & 0 & 0 & 0 & 0 & 0 & 0 & 0 \\
\textrm{female}       & 0 & 0 & 0 & 0 & 0 & 0 & 0 & 0 & 0 & 0 & 0 \\
\end{block}
\end{blockarray}\vspace*{-1.25\baselineskip}
\]

\[
\mathbf{M}_{officialLang} = 
\begin{blockarray}{lccccccccccc}
 & \textrm{UvL} & \textrm{AM} & \textrm{EM} & \textrm{Gy} & \textrm{Fr} & \textrm{EU} & \textrm{En} & \textrm{Ge} & \textrm{Fr} & \textrm{ma} & \textrm{fe} \\
\begin{block}{l[ccccccccccc]}
\textrm{U.v.d. Leyen} & 0 & 0 & 0 & 0 & 0 & 0 & 0 & 0 & 0 & 0 & 0 \\
\textrm{A. Merkel}    & 0 & 0 & 0 & 0 & 0 & 0 & 0 & 0 & 0 & 0 & 0 \\
\textrm{E. Macron}    & 0 & 0 & 0 & 0 & 0 & 0 & 0 & 0 & 0 & 0 & 0 \\
\textrm{Germany}      & 0 & 0 & 0 & 0 & 0 & 0 & 0 & 1 & 0 & 0 & 0 \\
\textrm{France}       & 0 & 0 & 0 & 0 & 0 & 0 & 0 & 0 & 1 & 0 & 0 \\
\textrm{EU}           & 0 & 0 & 0 & 0 & 0 & 0 & 0 & 0 & 0 & 0 & 0 \\
\textrm{English}      & 0 & 0 & 0 & 0 & 0 & 0 & 0 & 0 & 0 & 0 & 0 \\
\textrm{German}       & 0 & 0 & 0 & 0 & 0 & 0 & 0 & 0 & 0 & 0 & 0 \\
\textrm{French}       & 0 & 0 & 0 & 0 & 0 & 0 & 0 & 0 & 0 & 0 & 0 \\
\textrm{male}         & 0 & 0 & 0 & 0 & 0 & 0 & 0 & 0 & 0 & 0 & 0 \\
\textrm{female}       & 0 & 0 & 0 & 0 & 0 & 0 & 0 & 0 & 0 & 0 & 0 \\
\end{block}
\end{blockarray}\vspace*{-1.25\baselineskip}
\]

The product of these matrices is:
\[
\mathbf{M}_{\text{body}} = 
\begin{blockarray}{lccccccccccc}
 & \textrm{UvL} & \textrm{AM} & \textrm{EM} & \textrm{Gy} & \textrm{Fr} & \textrm{EU} & \textrm{En} & \textrm{Ge} & \textrm{Fr} & \textrm{ma} & \textrm{fe} \\
\begin{block}{l[ccccccccccc]}
\textrm{U.v.d. Leyen} & 0 & 0 & 0 & 0 & 0 & 0 & 0 & 1 & 0 & 0 & 0 \\
\textrm{A. Merkel}    & 0 & 0 & 0 & 0 & 0 & 0 & 0 & 1 & 0 & 0 & 0 \\
\textrm{E. Macron}    & 0 & 0 & 0 & 0 & 0 & 0 & 0 & 0 & 1 & 0 & 0 \\
\textrm{Germany}      & 0 & 0 & 0 & 0 & 0 & 0 & 0 & 0 & 0 & 0 & 0 \\
\textrm{France}       & 0 & 0 & 0 & 0 & 0 & 0 & 0 & 0 & 0 & 0 & 0 \\
\textrm{EU}           & 0 & 0 & 0 & 0 & 0 & 0 & 0 & 0 & 0 & 0 & 0 \\
\textrm{English}      & 0 & 0 & 0 & 0 & 0 & 0 & 0 & 0 & 0 & 0 & 0 \\
\textrm{German}       & 0 & 0 & 0 & 0 & 0 & 0 & 0 & 0 & 0 & 0 & 0 \\
\textrm{French}       & 0 & 0 & 0 & 0 & 0 & 0 & 0 & 0 & 0 & 0 & 0 \\
\textrm{male}         & 0 & 0 & 0 & 0 & 0 & 0 & 0 & 0 & 0 & 0 & 0 \\
\textrm{female}       & 0 & 0 & 0 & 0 & 0 & 0 & 0 & 0 & 0 & 0 & 0 \\
\end{block}
\end{blockarray}\vspace*{-1.25\baselineskip}
\]
This matrix $\mathbf{M}_{\text{P1P2}}$ shows which entities satisfy the body of the rule. Note that in our example, all elements are 0 or 1. If any matrix element would be greater than 1, such value would be replaced by 1. 

In this case, we see that the body of the rule is matched by paths $path_1$=\emph{\{birthCountry(U.v.d. Leyen, Germany), officialLang(Germany, German)\}}, $path_2$=\emph{\{birthCountry(Merkel, Germany), officialLang(Germany, German)\}} and $path_3$=\emph{\{birthCountry(E. Macron, France), officialLang(France, French)\}}. These paths correspond to the instantiations of ${\# \sigma_H: \sigma_H(\textbf{B}) \Vdash \kg}$ from Eq.~\ref{eq:stdconf}.

Now, we want to compare how this connects with the adjacency matrix for the head relation \emph{speaks}:
\[
\mathbf{M}_{head} = 
\begin{blockarray}{lccccccccccc}
 & \textrm{UvL} & \textrm{AM} & \textrm{EM} & \textrm{Gy} & \textrm{Fr} & \textrm{EU} & \textrm{En} & \textrm{Ge} & \textrm{Fr} & \textrm{ma} & \textrm{fe} \\
\begin{block}{l[ccccccccccc]}
\textrm{U.v.d. Leyen} & 0 & 0 & 0 & 0 & 0 & 0 & 1 & 1 & 0 & 0 & 0 \\
\textrm{A. Merkel}    & 0 & 0 & 0 & 0 & 0 & 0 & 0 & 0 & 0 & 0 & 0 \\
\textrm{E. Macron}    & 0 & 0 & 0 & 0 & 0 & 0 & 0 & 0 & 1 & 0 & 0 \\
\textrm{Germany}      & 0 & 0 & 0 & 0 & 0 & 0 & 0 & 0 & 0 & 0 & 0 \\
\textrm{France}       & 0 & 0 & 0 & 0 & 0 & 0 & 0 & 0 & 0 & 0 & 0 \\
\textrm{EU}           & 0 & 0 & 0 & 0 & 0 & 0 & 0 & 0 & 0 & 0 & 0 \\
\textrm{English}      & 0 & 0 & 0 & 0 & 0 & 0 & 0 & 0 & 0 & 0 & 0 \\
\textrm{German}       & 0 & 0 & 0 & 0 & 0 & 0 & 0 & 0 & 0 & 0 & 0 \\
\textrm{French}       & 0 & 0 & 0 & 0 & 0 & 0 & 0 & 0 & 0 & 0 & 0 \\
\textrm{male}         & 0 & 0 & 0 & 0 & 0 & 0 & 0 & 0 & 0 & 0 & 0 \\
\textrm{female}       & 0 & 0 & 0 & 0 & 0 & 0 & 0 & 0 & 0 & 0 & 0 \\
\end{block}
\end{blockarray}\vspace*{-1.25\baselineskip}
\]
Here, we can see that $\mathbf{M}_{head}$ and $\mathbf{M}_{body}$ overlap in two facts that relate to the head predicate: \emph{speaks(U.v.d Leyen, German)}, which connects to body $path_1$ and \emph{speaks(Macron, French)}, which connects to body $path_3$.
Hence, we got two complete instantiations for the rule $(\textbf{B} \Rightarrow H)$. According to Eq.~\ref{eq:support}, the value of support is 2.  

In this way, in the case of RLvLR, we demonstrated how matrix operations can be used to compute rule quality measures.

\subsection{Large Language Models for Learning Rules} 
This section discusses some of the recent studies using Large Language Models for learning rules.

\subsubsection{Hypotheses-to-Theories}

One of the recent framework, Hypotheses-to-Theories (HtT)~\cite{corr/abs-2310-07064} is designed to learn a set of rules from training examples, which is then used for reasoning over test samples using Large Language Models (LLMs). The framework is designed with the objective to target the issue of incorrect rule generation by LLMs which is often the case when LLMs rely on their implicit knowledge for rule creation instead of taking into account the problem at hand or the data provided. The framework employs both inductive and deductive reasoning through few-shot prompting. Inductive reasoning involves deriving general rules from specific observations, while deductive reasoning involves deriving new facts based on the existing ones. 

\paragraph{Induction Stage: Learning a Rule Library.}

The induction stage aims to learn rules from training examples without explicit rule annotations. For each training example (a question-answer pair), HtT prompts an LLM to generate rules for answering the question. Regular expressions are then used to extract rules from the LLM's output. Given the noisy nature of LLM reasoning, rules and accuracy metrics are collected from a sufficient number of training examples. The rules are filtered based on criteria from ~\cite{galarraga_amie_2013}, considering both coverage and confidence. Coverage indicates how likely a rule is to be reused, while confidence indicates how likely it is to be correct.

\paragraph{Deduction Stage: Reasoning using the Rule Library.}

The rule library generated in the induction phase is used for deductive reasoning prompt based on Chain-of-Thought prompting. The examples are modified to teach the LLM to retrieve rules from the library whenever it needs to generate a rule. If all the rules required by a question are present in the library, the LLM should generate correct rules for each step without errors. In order to facilitate the rule retrieval process the rule library is organized into a hierarchy using XML where each tag refers to a cluster.

\subsubsection{ChatRule}

ChatRule~\cite{corr/abs-2309-01538} is a framework designed for mining logical rules for Knowledge Graph (KG) reasoning tasks. The initial step involves an LLM-based rule generator that leverages both the semantic and structural information of KGs to prompt LLMs to create logical rules. To achieve this, the \textbf{rule sampler} conducts a Breadth-First Search (BFS) to sample several closed paths from KGs.


For instance, given a triple $(h_1, r_j, t_1)$, the closed-path is defined as a sequence of relations $r_1,\dots, r_n$ that connects $h_1$ and $t_1$ in KGs, i.e., $h_1 \stackrel{r_1}{\longrightarrow} h_2 \stackrel{r_2}{\longrightarrow} \dots \stackrel{r_j}{\longrightarrow} e_L$. For example, given a triple (Alice, GrandMother, Charlie), a closed-path p can be found as:

$$p := Alice \stackrel{Mother}{\longrightarrow} Bob \stackrel{Father}{\longrightarrow} Charlie,$$

which completes the triple (Alice, GrandMother, Charlie) in KGs. For a given target relation, a set of seed triples is selected from KGs, and BFS is conducted to sample a set of closed paths with lengths less than $L$, forming a set of rule instances. The actual entities in these rule instances are then replaced with variables to create rule samples. Each generated rule is verbalized into a natural language sentence, which is then incorporated into the prompt template.

To refine the generated rules, a rule ranking module assesses their quality by integrating facts from existing KGs. This ranking process utilizes support, coverage, confidence, and PCA confidence, as inspired by ~\cite{galarraga_amie_2013}. The ranked rules are then used for reasoning over KGs, addressing downstream tasks such as knowledge graph completion. In these tasks, candidate answers are ranked based on scores derived from coverage, confidence, or PCA confidence.


\section{Conclusions}
In this chapter, we described rule learning algorithms for knowledge graphs. We began with a comprehensive overview of the rule mining problem, followed by a discussion of popular rule mining algorithms to establish a foundation. Finally, we explored state-of-the-art neuro-symbolic rule learning approaches.

\section*{Acknowledgments}
Part of section \ref{sec:embeddings-rules} is adapted from the dissertation thesis of VZ \cite{zemanthesis}.

\bibliographystyle{vancouver}



\end{document}